\documentclass[a4paper,twoside]{article}

\usepackage{epsfig}
\usepackage{subcaption}
\usepackage{calc}
\usepackage{array}
\usepackage{amssymb}
\usepackage{amstext}
\usepackage{amsmath}
\usepackage{amsthm}
\usepackage{multicol}
\usepackage{pslatex}
\usepackage{apalike}
\usepackage{tabularx}
\usepackage{SCITEPRESS} 
\usepackage{hyperref}
\hypersetup{colorlinks,allcolors=black}
\begin{document}

\title{An Effective Deep Network for Head Pose Estimation without Keypoints}

\author{\authorname{Chien Thai, Viet Tran, Minh Bui, Huong Ninh and Hai Tran}
\affiliation{Computer Vision Department, Optoelectronics Center, Viettel Aerospace Institute, Vietnam}
\email{\{chientv13, vietth5, minhbq6, huongnt382, haitt27\}@viettel.com.vn}
}


\abstract{Human head pose estimation is an essential problem in facial analysis in recent years that has a lot of computer vision applications such as gaze estimation, virtual reality, driver assistance. Because of the importance of the head pose estimation problem, it is necessary to design a compact model to resolve this task in order to reduce the computational cost when deploying on facial analysis-based applications such as large camera surveillance systems, AI cameras while maintaining accuracy. In this work, we propose a lightweight model that effectively addresses the head pose estimation problem. Our approach has two main steps. 1) We first train many teacher models on the synthesis dataset - 300W-LPA to get the head pose pseudo labels. 2) We design an architecture with the ResNet18 backbone and train our proposed model with the ensemble of these pseudo labels via the knowledge distillation process. To evaluate the effectiveness of our model, we use AFLW-2000 and BIWI - two real-world head pose datasets. Experimental results show that our proposed model significantly improves the accuracy in comparison with the state-of-the-art head pose estimation methods. Furthermore, our model has the real-time speed of $\sim$300 FPS when inferring on Tesla V100.}

\onecolumn \maketitle \normalsize \setcounter{footnote}{0} \vfill

\section{\uppercase{Introduction}}
\label{sec:introduction}

Head pose estimation (HPE) is an important problem in facial analysis that has been extensively researched in recent years. Its application can be widely observed in lots of intelligent computer vision systems including virtual reality \cite{kumar2017kepler}, driver assistance \cite{schwarz2017driveahead,murphy2007head}, gaze estimation \cite{murphy2008head}, human-computer interaction \cite{seemann2004head,wang2019deep} and smart city surveillance.

The objective of head pose estimation is to accurately identify the orientation of heads of individuals found in images. Existing methods to solve this problem can be divided into two primary categories: landmark-based approaches \cite{cao2014face,lathuiliere2017deep,fanelli2011real,xiong2015global,sun2013deep,xin2021eva,bulat2017far,dementhon1995model} and landmark-free approach \cite{ruiz2018fine,yang2019fsa,zhou2020whenet,chang2017faceposenet}. Landmark-based methods use facial keypoints extracted by landmark detectors to regress the head pose angle. Recently, these approaches have achieved remarkable results since the usage of deep neural networks has greatly enhanced the quality of landmark detectors. However, the problem remains challenging due to the fact that not only a minor error of landmark detectors may adversely affect the head pose estimation but learning the relation between the geometric distribution of facial landmarks and head poses is not a trivial task. Furthermore, using landmark detection as a preprocessing step imposes a computational burden for the whole process of estimating head angle which hinders its usage for real-time applications. Landmark-free methods, on the other hand, directly predict the head poses from images without detecting facial keypoints which results in their fast execution time.

In addition to the above approaches, some works utilize depth information from depth cameras \cite{meyer2015robust,fanelli2011real,mukherjee2015deep,martin2014real}. Although this approach provides a prominent result, it still has some limitations. The depth cameras are sensitive to illumination change and light conditions so that they often yield substandard results in an uncontrolled environment. Moreover, they are very expensive and use more storage and transfer time, so they are often impractical for real-time applications.

Because of the importance of the head pose estimation problem and in order to minimize the processing time of the model when deploying on large systems or embedded platforms, our goal is to design a lightweight architecture that solves this task while still guaranteeing remarkable performance. For having a compact and simple model, our network uses ResNet18 architecture as a backbone. The contributions of our work can be summarized as follows:
\begin{itemize}
\item We address a major mistake found in HopeNet \cite{ruiz2018fine} in which annotated face boxes are mislabeled. We prove that correcting those mislabeled boxes can significantly improve the accuracy of the head pose estimation task.
\item An end-to-end deep architecture designed to solve head pose estimation problem is proposed. A lightweight model is trained to this task via the knowledge distillation process.
\item Experiments are conducted to evaluate the performance of our method on two challenging head pose datasets (BIWI and AFLW-2000). Our method achieves state-of-the-art performance when evaluating on the head pose dataset.
\end{itemize}

The rest of the paper is organized as follows: Section 2 puts forward some related works on head pose estimation problem. In section 3, we present our proposed method. Section 4 discusses the datasets, experiments, results, and ablation study. Finally, the conclusion and future work are discussed in Section 5.

\section{\uppercase{Related Works}}
\textbf{Convolutional neural networks} (CNNs) are widely used in computer vision tasks and gradually replace the traditional image processing methods. CNN is designed to automatically learn the spatial features of the image by using convolution kernels. With many convolutional layers, deep networks can extract high-level semantic features. He et al. \cite{he2016deep} propose the Residual Network to train the much deeper convolutional neural network. ResNet uses a skip connection between the current layer and the previous layer which can learn the identity mapping and solve the vanishing gradient problem. Because of its powerful and simple architecture, ResNet and its variants \cite{xie2017aggregated,zhang2020resnest,gao2019res2net} are widely used in many computer vision applications and deliver high performance.

\textbf{Human head pose estimation} has been researched over the past 25 years with many different approaches. Appearance Template \cite{niyogi1996example,beymer1994face,sherrah2001face,ng2002composite,sherrah1999understanding} is the method that compares the input image with a set of labeled templates and assigns it to the most similar template. Detector arrays \cite{huang1998face,zhang2006head,jones2003fast} estimate head pose by training multiple face detectors for the different discrete poses. 

Many approaches are based on facial landmarks from the input image to estimate the head pose. With the progress of landmarks detection, landmark-based methods demonstrate superior performance. Dementhon et al. \cite{dementhon1995model} proposed Pose from Orthography and Scaling with Iterations which determines the head pose by 3D computer vision techniques for the given 2D face landmarks. FAN \cite{bulat2017far} using deep neural network to estimate 3D face models. EVA-GCN \cite{xin2021eva} constructs a landmark-connection graph and leverages the Graph Convolution Network \cite{yan2018spatial} to learn the nonlinear relationships between head poses and distribution of facial keypoints.

Multi-task methods combine the head pose estimation problem with other related facial analysis problems, such as face detection, keypoints detection. Some works show that learning with related tasks yields better results than learning individual tasks independently \cite{chen2014joint,kumar2017kepler,zhu2012face,ranjan2017all}. KEPLER \cite{kumar2017kepler} predicts face detection and pose estimation jointly by using Heatmap-CNN to capture structured global and local features. Hyperface \cite{ranjan2017hyperface} presents a convolutional neural network for simultaneous face detection, landmarks localization, pose estimation, and gender recognition.

Gu et al. \cite{gu2017dynamic} proposed a dynamic facial analysis that uses a recurrent neural network. They improve head pose estimation and facial landmarks localization by leveraging the time dimension from videos instead of a single frame.

For accurate head pose estimation, some methods utilize 3D information of depth images. Meyer et al. \cite{meyer2015robust} perform head pose estimation by registering 3D morphable models to depth images, using the particle swarm optimization and the iterative closest point algorithm. Fanelli et al. \cite{fanelli2011real} using Random Regression Forests to regress the head pose estimation of depth images. 

\begin{figure*}[htp]
    \centering
    \includegraphics[scale=0.65]{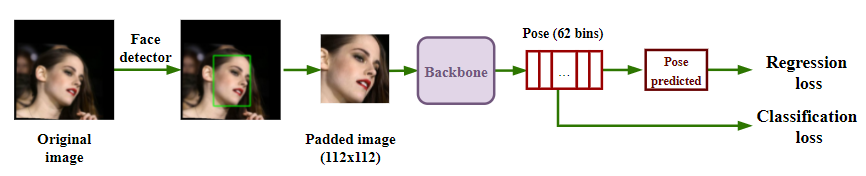}
    \caption{The overview of the head pose model. The original image is passed through the face detector to get the bounding box of the objective face. The detected face is padded to a squared image and resized to 112x112. The head pose model extract 62 dimensions distribution vector for the given image. The predicted pose is calculated by the expectation of this vector. For each Euler angle, the classification loss is the cross-entropy loss between distribution vector and one-hot vector, the regression loss is the mean square error of ground truth and predicted pose.}
    \label{fig:galaxy}
\end{figure*}
Recent works directly predict the Euler angles from a single RGB image by using a deep neural network and achieve prominent performance. HopeNet \cite{ruiz2018fine} proposed a multi-loss framework that combines binned pose classification and regression loss for each Euler angle. By using a very stable softmax layer and cross-entropy for binned classification loss, the network obtained robust neighborhood prediction of the head pose. FSA-Net \cite{yang2019fsa} employs the soft stagewise regression scheme by training classification and regression objectives of the features from multiple stages. It provides a compact model and accurate prediction. WHENet \cite{zhou2020whenet} proposed wrapped loss to estimate the full 360-degree range of yaw angle. Our proposed network has similar architecture to HopeNet \cite{ruiz2018fine}, but has a smaller model size and achieves better performance on two challenging head pose datasets - BIWI and AFLW-2000.

\section{\uppercase{Proposed Method}}
In this section, we describe the major disadvantage of previous work and the method to mitigate this problem. After that, we explain the proposed method to construct an effective head pose estimation model via knowledge distillation process.

The head pose estimation problem can be mathematically formulated as: Given a set of training images \textit{$X = \{x^i | i = 1..N\}$} and ground truth \textit{$Y = \{y^i | i = 1..N\}$}, where \textit{N} is number of images, and \textit{$y^i$} is 3D vector of image \textit{$x^i$} corresponding to three Euler angles (yaw, pitch, roll), the goal is to find a function \textit{F} so that the absolute difference between \textit{F(x)} and the real head pose \textit{y} for the given image \textit{x} as small as possible.

Inspired by HopeNet \cite{ruiz2018fine}, we design a network using a multi-loss framework to solve this problem. HopeNet casts the regression problem of head pose estimation as a classification problem by dividing the poses range into 66 bins, each bin contains 3 units of degree. The predicted pose is the expected value of classes distribution.

Further investigating  HopeNet, we found that it is the preprocessing data that hinders its performance. They loosely crop around the bounding box of a face on the image and resize the cropped image to 224x224 before fitting the model. Because the height of the face bounding box is often longer than the width, it slightly changes the real head pose and causes a negative effect on the training and testing phase.

To mitigate this problem, we padded the bounding box of face to squared shape. Given a bounding box ($x_{1}$, $y_{1}$, $x_{2}$, $y_{2}$), the padding size \textit{k} is calculated by $|x_{2}$ – $x_{1}$ – $y_{2}$ + $y_{1}|$ (the absolute difference between width and height). If the height $h$ = $x_{2}$ – $x_{1}$ is longer than the width $w$ = $y_{2}$ – $y_{1}$, the new coordinates of bounding box ($x'_{1}$, $y'_{1}$, $x'_{2}$, $y'_{2}$) are:
\begin{align*}
    x'_{1} &= x_{1} \\
    x'_{2} &= x_{2} \\
    y'_{1} &= y_{1} - [k/2] \\
    y'_{2} &= y_{2} + [k/2]
\end{align*}
and vice versa. After getting the square image of faces, we resize it to (112, 112) in order to decrease the computation cost when training and inference.

Unlike HopeNet, we divide the poses range from -93 to 93 into 62 bins for each Euler angle. The classification loss of angle is cross-entropy loss between softmax output of model and pose’s corresponding one-hot vector:
\begin{equation}
    \mathcal{L}_{cls}^{angle} = \sum_{i = 1}^{N} y'^{i}*log(\hat{y}^{i})
\end{equation}
where $y'^{i}$ and $\hat{y}^{i}$ are respectively one-hot vector of pose and predicted softmax output for given input $x^{i}$.

The predicted pose of $x^{i}$ is expected values of softmax output that is denoted by $r^{i}$. The regression loss of angle is mean squared error between the ground truth labels $y^{i}$ and the predicted pose $r^{i}$:
\begin{equation}
    \mathcal{L}_{reg}^{angle} = \frac{1}{N}\sum_{i = 1}^{N} \|r^{i} - y^{i}\|^2
\end{equation}

\begin{figure*}[htp]
    \centering
    \includegraphics[scale=0.35]{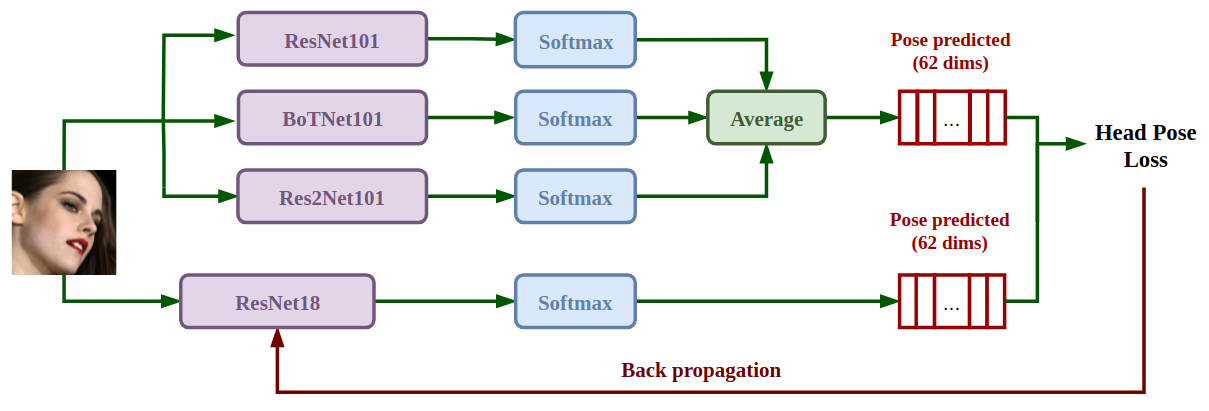}
    \caption{The overview of proposed method. The student model using ResNet18 backbone. The head pose loss is the sum of Kullback-Leibler Divergence loss between softmax output of student model and ensemble output of head pose teacher models on each of yaw, pitch, roll angle. The total loss is sum of distillation loss of three Euler angles}
    \label{fig:galaxy}
\end{figure*}

The total loss is composed by three separate losses, each loss is calculated by the sum of classification and regression loss of angle, as following:
\begin{equation}
    \mathcal{L} = \sum_{angle} \mathcal{L}_{cls}^{angle} + \mathcal{L}_{reg}^{angle}
\end{equation}
where \textit{$angle \in \{yaw, pitch, roll\}$}

The above method uses hard labels to train head pose estimation models. Inspired by \cite{hinton2015distilling}, we use knowledge distillation to construct a compact model while enhancing the performance of this task. Our network uses ResNet18 \cite{he2016deep} as the backbone, a simple and small architecture which is trained to match the output of head pose teacher models (pseudo label). With supervised learning, models are trained to match the same labels but with the different initiation and architectures, they will focus on distinctive features. So, we ensemble outputs of several strong head pose models to get more informative teacher features.

Given \textit{$N_{teacher}$} head pose models, we ensemble by calculating mean regression outputs of them. It is equal to the expected value of mean softmax outputs of these models. So, the output after ensemble n teacher head pose models is:
\begin{equation}
    y_{i}^{ens} = \frac{1}{N_{teacher}}\sum_{j=1}^{N_{teacher}} \hat{y}^{j}_{i} 
\end{equation}
where $\hat{y}^{j}_{i}$ is softmax ouput of head pose teacher model \textit{j} for given image $x^{i}$

The loss function for head pose task is Kullback-Leibler Divergence between softmax output of student model $\hat{y}^{t}$ and output ensemble of n teacher models $y^{ens}$:
\begin{equation}
    \mathcal{L}_{headpose} = -\sum_{i=1}^{N} y^{ens}_{i}*log(\frac{\hat{y}^{t}_{i}}{y^{ens}_{i}})
\end{equation}

Because head pose estimation is a challenging task, we found that the stronger model with a lot of parameters and computation cost, the more model's capacity to achieve good results. Base on the performance on ImageNet dataset \cite{deng2009imagenet}, we train three head pose teacher models from scratch whose backbones are chosen respectively as ResNet101 \cite{he2016deep}, BotNet101 \cite{srinivas2021bottleneck}, and Res2Net101 \cite{gao2019res2net}. After that, we train a head pose model with backbone ResNet18 by the aforementioned head pose knowledge distillation strategy. 

In our experiment, we observed that the big models (teacher models) often give larger probabilities to the bins in the proximity of the truth bin and smaller scores to the ones far away. This is valuable information (i.e. the faces in a bin are more likely the faces in its neighbor bins) but it has very little effect on the cross entropy cost function during training if the probabilities are so close to zero. This means the soft targets of the teacher models attain a variety of information than one-hot labels, which helps the small model (student model) learn easily. So, we argue that the distilled model can preserve the generalization of the teacher models and reaches highly accurate results.

\begin{table*}[htbp]
\centering
\renewcommand{\arraystretch}{1.4}
\begin{center}
\caption{Mean average error of Euler angles across both state-of-the-art landmark-based and landmark-free methods on the BIWI and AFLW2000 dataset}
\begin{tabularx}{\textwidth}{m{5.5cm} | >{\centering\arraybackslash}X  >{\centering\arraybackslash}X >{\centering\arraybackslash}X >{\centering\arraybackslash}X | >{\centering\arraybackslash}X >{\centering\arraybackslash}X >{\centering\arraybackslash}X >{\centering\arraybackslash}X  }
    \hline
     & & \textbf{BIWI} & & & & \textbf{AFLW-2000} & & \\
    \hline 
   \textbf{Model} & \textbf{Yaw} & \textbf{Pitch} & \textbf{Roll} & \textbf{MAE} & \textbf{Yaw} & \textbf{Pitch} & \textbf{Roll} & \textbf{MAE}\\
   \hline
   KEPLER \cite{kumar2017kepler}  & 8.80  & 17.3 & 16.2 & 13.9 & - & - & - & -\\
   \hline
   FAN \cite{bulat2017far} & 8.53  & 7.48 & 7.63 & 7.89 & 6.36  & 12.3 & 8.71 & 9.12 \\
   \hline
   Dlib \cite{kazemi2014one} &  16.8 & 13.8 & 6.19 & 12.2 &  23.1 & 13.6 & 10.5 & 15.8 \\
   \hline
   3DDFA \cite{zhu2016face}  & - & - & - & - & 5.40  & 8.53 & 8.25 & 7.39 \\
   \hline
   EVA-GCN \cite{xin2021eva} & 4.01 & 4.78 & 2.98 & 3.92 & 4.46 & 5.34 & 4.11 & 4.64 \\
   \hline
   HopeNet ($\alpha = 2$) \cite{ruiz2018fine} & 5.17 & 6.98 & 3.39 & 5.18 & 6.47 & 6.56 & 5.44 & 6.16 \\
   \hline
   HopeNet ($\alpha = 1$) \cite{ruiz2018fine}& 4.81 & 6.61 & 3.27 & 4.90 & 6.92 & 6.64 & 5.67 & 6.41 \\
   \hline
   SSR-Net-MD \cite{yang2018ssr}& 4.49 & 6.31 & 3.61 & 4.65 & 5.14 & 7.09 & 5.89 & 6.01\\
   \hline
   FSA-Caps-Fusion \cite{yang2019fsa} & 4.27 & 4.96 & 2.76 & 4.00 & 4.50 & 6.08 & 4.64 & 5.07 \\
   \hline
   WHENet-V \cite{zhou2020whenet}& \textbf{3.60} & 4.10 & 2.73 & 3.48 & 4.44 & 5.75 & 4.31 & 4.83 \\
   \hline
   \textbf{EHPNet (Ours)} & 3.68 & \textbf{4.03} & \textbf{2.57} & \textbf{3.43} & \textbf{3.23} & \textbf{5.54} & \textbf{3.88} & \textbf{4.15}\\
   \hline
\end{tabularx}
\renewcommand{\arraystretch}{1}
\end{center}
\end{table*}

\section{Experimental results}
In this section, we describe the datasets for training and testing, implementation, results, comparisons with other state-of-the-art methods and the ablation study.
\subsection{Dataset}
\begin{figure}[htp]
    \centering
    \includegraphics[scale=0.5]{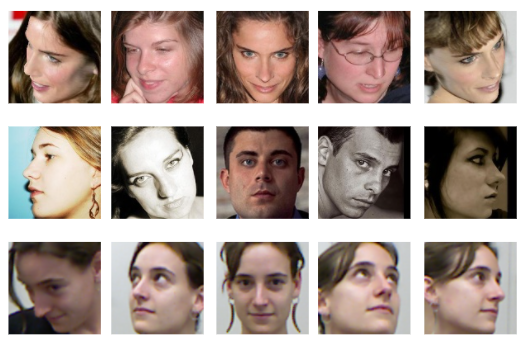}
    \caption{Some examples of face image from the datasets. The first row is from the 300W-LPA \cite{hsu2019edge} which is a synthetically dataset. The second row and third row are respectively from the AFLW-2000 \cite{zhu2016face} and BIWI \cite{fanelli2011real} - two real-world datasets}
    \label{fig:galaxy}
\end{figure}

\textbf{Headpose dataset:} In our experiment, we use three popular datasets for the head pose estimation problem: 300W-LPA \cite{hsu2019edge}, AFLW-2000 \cite{zhu2012face}, and BIWI \cite{fanelli2011real} datasets. 300W-LPA is a synthetically expanded dataset that provides over 350000 images across large poses. The AFLW-2000 dataset provides head pose ground truth and corresponds to 68 landmark points among 2000 3-D face images. Images in the AFLW-2000 dataset have large pose annotation and various lighting conditions. BIWI dataset uses a Kinect v2 device to record RGB-D video of different subjects. It contains 24 videos of 20 subjects across different head poses. There are roughly 15000 samples in this dataset, each sample contains RGB and depth images, and pose annotations were created by using depth information.

Following HopeNet, we use 300-LPA for training while testing on AFLW-2000 and BIWI - two real-world datasets. In our case, we only use RGB images for training in these datasets. We run RetinaFace \cite{deng2019retinaface} on all images to get the coordinate of the bounding box of faces.

\subsection{Implementation}
For better estimating on low-resolution face images, we augment the head pose training dataset by random downsampling and upsampling to original image size, randomly adjust brightness and contrast, blur by random Gaussian kernel.
We randomly flip the image and relabel the yaw and roll angle of the flip image to -yaw and -roll to get more training data. 

We use Pytorch for implementing the proposed network. We use 100 epochs to train the teacher networks with hard labels and 200 epochs for the knowledge distillation process. The chosen optimizer is Adam with initial learning-rate is 1e-4. The learning rate is reduced by the cosine annealing strategy. The experiments are performed on a computer with a Tesla V100 GPU.
\begin{figure*}[htbp]
    \centering
    \includegraphics[scale=0.57]{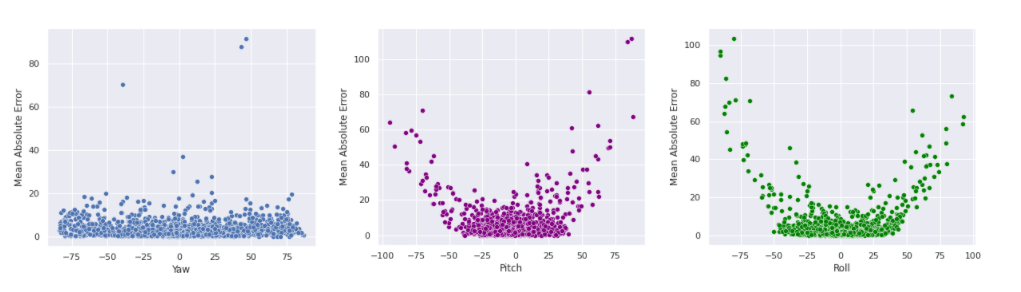}
    \caption{The scatter plot between yaw, pitch, roll values and errors on AFLW-2000 dataset}
    \label{fig:galaxy}
\end{figure*}
\begin{figure*}[htbp]
    \centering
    \includegraphics[scale=0.57]{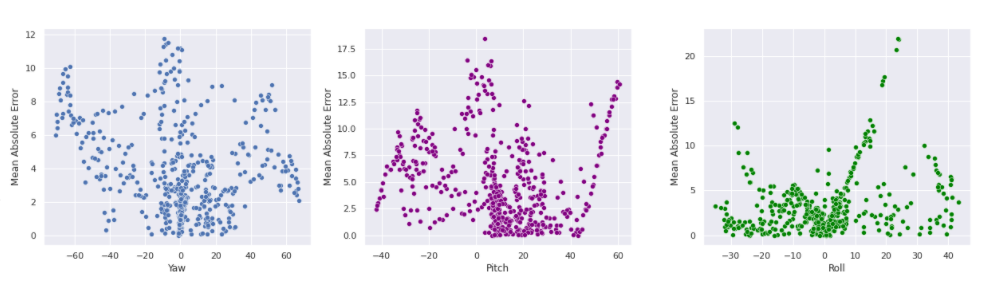}
    \caption{The scatter plot between yaw, pitch, roll values and errors on BIWI Dataset}
    \label{fig:galaxy}
\end{figure*}
\subsection{Results}

We compare our proposed network with other state-of-the-art head pose estimation methods on BIWI and AFLW datasets. KEPLER \cite{kumar2017kepler}, FAN \cite{bulat2017far}, Dlib \cite{kazemi2014one} and EVA-GCN \cite{xin2021eva} are landmark-based methods. KEPLER \cite{kumar2017kepler} uses a modified GoogleNet to predict facial landmark points and pose at the same time. Dlib \cite{kazemi2014one} is a face library that uses an ensemble of regression trees to detect landmarks. FAN \cite{bulat2017far} is a state-of-the-art landmark detection method. EVA-GCN \cite{xin2021eva} is a state-of-the-art landmark-based method which constructs a landmark-connection graph and leverages the Graph Convolution Network \cite{yan2018spatial} to learn the nonlinear relationships between head poses and distribution of facial keypoints. HopeNet \cite{ruiz2018fine}, FSANet \cite{yang2019fsa} and WHENet \cite{zhou2020whenet} are landmark free methods which treat the regression problem as classification problem by dividing the poses range to different classes. The $\alpha$ coefficient of HopeNet is the weight of the regression losses. 

Table 1 shows the comparisons of our proposed network with these above models respectively on BIWI and AFLW-2000 datasets. The evaluation metric is the mean absolute error of Euler angles. 

As shown in Table 1, our proposed EHPNet achieves state-of-the-arts on both AFLW-2000 and BIWI datasets. It outperforms the previous state-of-the-art WHENet by 14.9\% and 1.15\%, respectively. EHPNet has a similar architecture with WHENet and HopeNet but uses a smaller model and image size. Even so, it has more significant improvement compared to WHENet and HopeNet. Furthermore, our model has a real-time speed of 300FPS when inferring on Tesla V100.

Figure 6 shows some visualization of predicted images on BIWI, AFLW-2000. Besides achieving good results on images with variant poses and various lighting conditions, the network can also predict well on face images with high occlusion.

\begin{figure}[htp]
    \centering
    \includegraphics[scale=0.35]{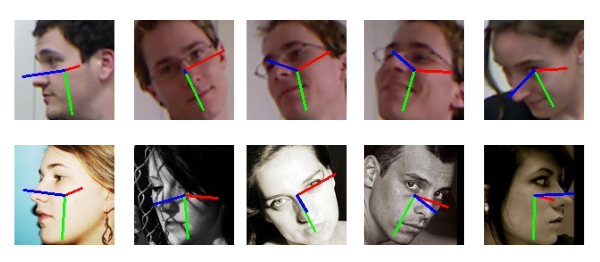}
    \caption{Results of the proposed network. The blue line indicates towards the front of the face, the green line pointing downward direction and the red line pointing to the side. The first row is the prediction on the BIWI dataset. The second row is the estimation result on images with various lighting conditions of AFLW-2000. 
}
    \label{fig:galaxy}
\end{figure}

\begin{table*}
\begin{center}
    \centering
    \renewcommand{\arraystretch}{1.5}
    \caption{The impact of different backbone and distillation training on head pose estimation models. The evaluation metric is the mean absolute error of Euler angles. }
    \begin{tabularx}{\textwidth}{m{3cm} | >{\centering\arraybackslash}X >{\centering\arraybackslash}X >{\centering\arraybackslash}X >{\centering\arraybackslash}X  | >{\centering\arraybackslash}X >{\centering\arraybackslash}X >{\centering\arraybackslash}X >{\centering\arraybackslash}X  }
    \hline
     & & \textbf{BIWI} & & & & \textbf{AFLW-2000} & & \\
     \hline 
     \textbf{Backbone} & \textbf{Yaw} & \textbf{Pitch} & \textbf{Roll} & \textbf{MAE} & \textbf{Yaw} & \textbf{Pitch} & \textbf{Roll} & \textbf{MAE} \\
     \hline
     ResNet18   & 3.969 & 4.849 & 2.869 & 3.897 & 3.785 & 5.642 & 4.238 & 4.555 \\
     \hline
     ResNet101   & 3.680 & 3.945 & 2.755 & 3.460 & 3.249 & 5.276 & 3.821 & 4.115 \\
     \hline
     BotNet101  & 3.876 & 4.066 & 2.528 & 3.489 & 3.559 & 5.109 & 3.697 & 4.135 \\
     \hline
     Res2Net101 & 3.827 & 3.939 & 2.669 & 3.478 & 3.223 & 5.080 & \textbf{3.556} & 3.953 \\
     \hline
     Ensemble  & 3.688 & \textbf{3.859} & \textbf{2.508} & \textbf{3.352} & \textbf{3.169} & \textbf{5.009} & 3.560 & \textbf{3.913} \\
     \hline
     Distilled ResNet18 & \textbf{3.683} & 4.033 & 2.571 & 3.429 & 3.226 & 5.345 & 3.876 & 4.148 \\
     \hline
    \end{tabularx}
    \renewcommand{\arraystretch}{1.25}
\end{center}
    \label{tab:my_label}
\end{table*}
The scatter diagrams in Figure 4 and Figure 5 show the influence of the pose's value on prediction results for each angle on the AFLW-2000 and BIWI datasets. On AFLW-2000, the yaw angle has a smaller mean absolute error than pitch and roll angles and has a stable prediction for pose range. For the pitch and roll angles, the model tends to predict well if the value of pose is as close to 0. As shown in Figure 4, there is some prediction that has a very big error. We find that this happens because the head poses on AFLW-2000 are provided by using 3D landmarks, so some examples can have a big difference in a pose if viewed as RGB images. On the BIWI dataset, the discrepancy between ground truth and prediction is not significant but the trending is slightly changed. The model predicts better on yaw and roll angle. The higher pose's value of pitch angle leads to face occlusion in the image and makes the model confused. As shown in Figure 5, many samples from the BIWI dataset don't follow the trend. For example, a sample which has a yaw angle value close to zero has a maximum error. In our experiments, we observed that, although it has a small value of the yaw angle, it has a large pitch and roll so the face can be occluded, and leads to wrong predictions.

\subsection{Ablation study}

We have conducted the ablation study when changing the backbone and using a pseudo label from teacher models for the head pose estimation task. As shown in Table 2, the result has significant improvement when using padding instead of resizing the cropped face image like HopeNet. By using ensemble, the mean absolute errors are slightly decreased on both BIWI and AFLW datasets. The small head pose model achieves better accuracy when training via the knowledge distillation process. With ResNet18 as a backbone, the pose model using pseudo labels from the output of ensemble many teacher models is better than the same model using the hard label. The distilled head pose model has equivalent results to its teacher, even better. 

In our experiment, we observed that these models can predict the same result for the yaw angle. But for pitch and roll angle, the complex head pose model works better. Among three teacher models, each one can predict better at a specific pose interval (i.e. model ResNet101 achieves the smallest error of yaw with 3.680, while Res2Net101 outperforms the two others with the pitch error of 3.939 and the last one BotNet101 attains the best roll error of 2.528), but worse at the others. By preserving the generalization of each model, the ensemble results have a stable prediction on the poses range. Overall, training the baseline model with hard targets leads to severe over-fitting, whereas training the same model with ensemble soft targets is able to recover better generalization and achieves competitive results.
\section{\uppercase{Conclusions}}
\label{sec:conclusion}
In this paper,  we have presented an EHPNet,  which can directly, accurately, and robustly predict the head rotation from a single RGB image.  This is achieved by mitigating the disadvantages of previous work, along with distilling the knowledge from many robust head pose estimation teacher models.  By using ResNet18 architecture as a backbone, the model is compact and usable in many computer vision applications. The proposed network outperforms both landmark-based and landmark-free methods and achieves state-of-the-art results on both the AFLW-2000 and BIWI datasets, with higher respectively 14.9\% and 1.15\% than WHENet. 

In the future, we would like to use a lower computation network as well as reduce the input image resolution. Besides, more effective knowledge distillation techniques will be used to help the student model achieve better accuracy.

\bibliographystyle{apalike}
% {\small
% \bibliography{example}}

\end{document}